\documentclass{article}

\usepackage{arxiv}

\usepackage[utf8]{inputenc} 
\usepackage[T1]{fontenc}    
\usepackage{hyperref}       
\usepackage{url}            
\usepackage{booktabs}       
\usepackage{amsfonts}       
\usepackage{amsmath}
\usepackage{nicefrac}       
\usepackage{microtype}      
\usepackage{lipsum}
\usepackage{graphicx}

\usepackage{multirow} 
\usepackage{makecell}
\newcounter{magicrownumbers}
\newcommand\rownumber{\stepcounter{magicrownumbers}\arabic{magicrownumbers}}

\usepackage{xcolor} 

\title{A Model-Free Sampling Method for Basins of Attraction Using Hybrid Active Learning (HAL)}

\author{
	Xue-She Wang, Samuel A. Moore, James D. Turner \& Brian P. Mann \\
	Dynamical Systems Research Laboratory \\
	Department of Mechanical Engineering \& Materials Science \\
	Duke University \\
	Durham, NC 27708, USA \\
	\texttt{sam.a.moore@duke.edu} \\
}

\begin{document}
\maketitle

\begin{abstract}
Understanding the basins of attraction (BoA) is often a paramount consideration for nonlinear systems. Most existing approaches to determining a high-resolution BoA require prior knowledge of the system's dynamical model (e.g., differential equation or point mapping for continuous systems, cell mapping for discrete systems, etc.), which allows derivation of approximate analytical solutions or parallel computing on a multi-core computer to find the BoA efficiently. However, these methods are typically impractical when the BoA must be determined experimentally or when the system’s model is unknown. This paper introduces a model-free sampling method for BoA. The proposed method is based upon hybrid active learning (HAL) and is designed to find and label the ``informative'' samples, which efficiently determine the boundary of BoA. It consists of three primary parts: 1)~additional sampling on trajectories (AST) to maximize the number of samples obtained from each simulation or experiment; 2)~an active learning (AL) algorithm to exploit the local boundary of BoA; and 3)~a density-based sampling (DBS) method to explore the global boundary of BoA. An example of estimating the BoA for a bistable nonlinear system is presented to show the high efficiency of our HAL sampling method.

\end{abstract}


\section{Introduction}

Equilibria and their stabilities play a fundamental role in dynamical systems. For nonlinear systems, basins of attraction (BoA) provide a useful mapping from initial conditions to stable equilibria states (attractors). In other words, if the system starts at any state in an attractor's basin, its trajectory will asymptotically converge to this attractor. An accurate estimation of BoA is of vital importance in analyzing a system's control stability or predicting its dynamic behavior.


In addition to several classic works on the characterization of BoA~\cite{genesio1985estimation, chiang1988stability, chiang1989stability, alberto2011characterization}, the estimate of BoA with Lyapunov functions (LFs) has been thoroughly investigated. Lyapunov-based methods compute a LF as a local stability certificate and its sublevel sets, in which the function decreases along the flow, provide invariant subsets of the BoA~\cite{margolis1963control}. Its contribution includes the use of maximal LFs~\cite{rozgonyi2010determining, vannelli1985maximal}, piecewise LFs~\cite{ohta2002piecewise, ohta1999stability}, logical compositions of LFs~\cite{balestrino2011logical}, and occupation measures~\cite{Henrion2013Convex}. Relatively new Lyapunov-based methods use SOS programming (an optimization technique based on sum of squares of polynomials) to optimize and enlarge the estimate of BoA. The use of SOS programming proved to work for BoA in polynomial systems~\cite{jarvis2003some, ichihara2009optimal, chesi2004computing}, robust BoA in uncertain polynomial systems~\cite{chesi2004estimating, chesi2011domain}, and BoA in non-polynomial systems~\cite{chesi2005domain, chesi2009estimating}.

Many non-Lyapunov methods to estimate the BoA have also been broadly investigated. For instance, cell-to-cell mapping considers the state space as a collection of discrete state cells to significantly improve the computational speed of sweeping the entire state space~\cite{hsu2013cell, hsu1980unravelling, hsu1981generalized}. Backward mapping determines a small sufficient BoA first and then enlarges it in a systematic manner~\cite{hsu1977determination}. There is also considerable work on tracking the manifolds that define the BoA boundaries directly. Trajectory reversing obtains the boundary of BoA by integrating backward from unstable equilibria~\cite{genesio1985estimation}. Continuation methods based on boundary-value problems were used to compute the global invariant manifolds of vector fields~\cite{krauskopf1999two, osinga2010investigating, osinga2014computing}. Set-oriented numerical methods based on multilevel subdivision procedures are useful tools for approximating various types of invariant sets or manifolds~\cite{dellnitz1997subdivision, dellnitz2002set, krauskopf2005survey}.

Although a large number of BoA estimation methods have been proposed for nonlinear systems, there are still several application scenarios where the existing methods work inefficiently or can not be implemented: 1) when a map describing the BoA is needed, and 2) when the system's dynamical model is unknown. First, in some reinforcement learning applications~\cite{wang2020constrained}, the attractor where the system will converge is evaluated in each time step given the instantaneous state (probably million or billion operations for entire learning process). A map describing BoA, which has an input of an arbitrary state and output of the corresponding attractor, is therefore needed for fast predicting a system's future behavior without long-term integration. Many existing methods, which are based on finding BoA boundaries or manifolds, can provide a useful visualization and qualitative analysis, but further information is required for quantitative results. 

In addition, basins of attraction oftentimes need to be estimated from experiments when the form of the model is either unknown or approximate, e.g.~the use of a linear friction model~\cite{Wang2018Dynamics} or a simplified magnet model~\cite{Wang2019Nonlinear}. Given that experiments give no prior knowledge of governing equations (e.g., differential equation or point mapping for continuous systems, cell mapping for discrete systems, etc.), the existing methods based on Lyapunov functions or backwards time integration become difficult to implement.  While  cell-to-cell mapping could still work theoretically, generating cell mappings from experimental data can be time-consuming. To elaborate, the ability to use parallel computing for multiple short-term simulations helps make this approach efficient. However, experimental data can typically only be obtained one experiment at a time. Dividing a long-term experiment into short-term ones oftentimes gives no benefit to efficiency, and can even be more time-consuming for spending much time on initializing experimental setups in each short-term experiment. 

This paper proposes a novel sampling method for BoA estimation that 1)~provides a continuous map from an arbitrary state to its resting attractor, and 2)~requires no prior knowledge of the system's underlying math model. It's assumed that the BoA boundary is not fractal and the number of attractors is finite. The map describing BoA can therefore be considered a classifier, identifying to which basin a state belongs. An alternative interpretation of this work is a data-efficient sampling method for training a BoA classifier. 


\begin{figure}
	\centering
	\includegraphics[width=0.8\linewidth]{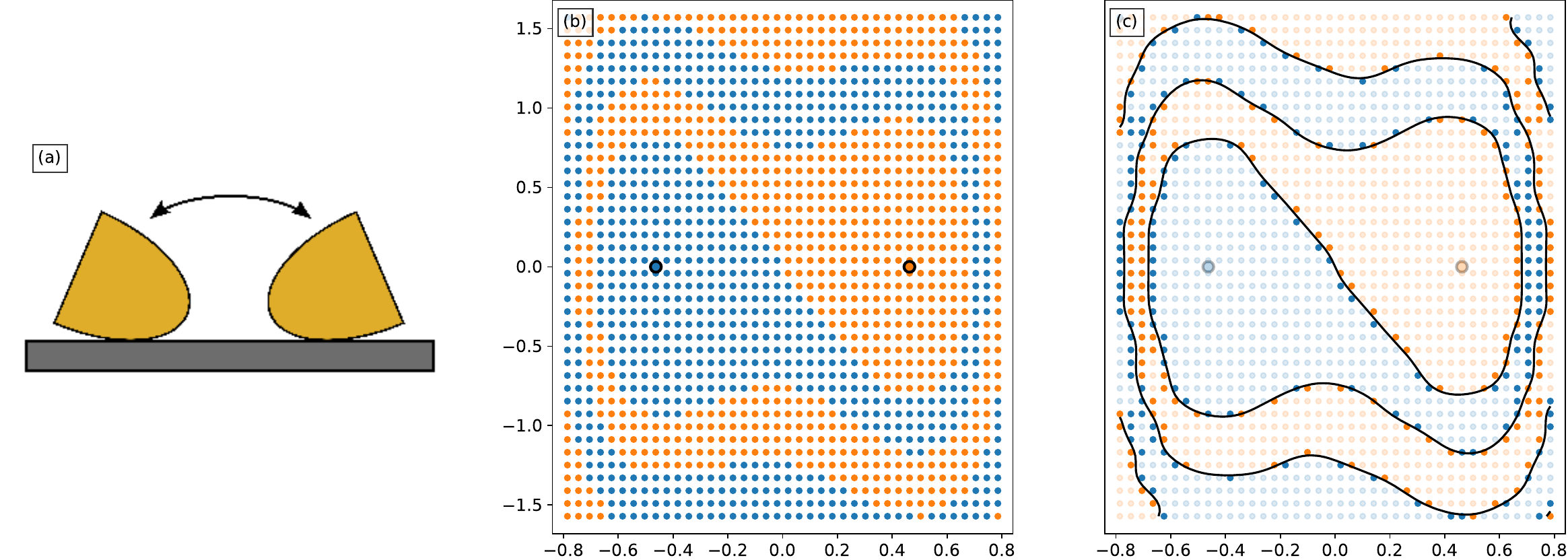}
	\caption{BoA for a bistable semi-elliptical rocking disk in~\cite{Wang2018Dynamics}. (a) Two stable equilibria (fixed-point attractor) of the rocking disk: left-tilt and right-tilt. (b) Uniform sampling of $40 \times 40$ initial conditions, where the blue and orange points represent the basin of the left-tilt and right-tilt equilibrium respectively. The points circled in black are the fixed-point stable equilibria. (c) The black line represents the BoA boundary, which is determined by only the highlighted samples ($20\%$). The majority of samples ($80\%$) are far away from the boundary and have no effect on estimating the BoA.}
	\label{fig:UFS}
\end{figure}

\section{Sampling Method}
\label{sec:method}

Uniform sampling is a brute-force method to obtain the training data. For the example in Fig.~\ref{fig:UFS}(a), a bistable rocking disk has two equilibria (fixed-point attractors) of rest angles: left-tilt and right-tilt. Its initial angle and angular velocity determine the equilibrium resting angle where it will eventually settle down, thus comprising the system's BoA. As shown in Fig.~\ref{fig:UFS}(b), the uniform sampling method divides the state domain into, e.g., a $40 \times 40$ grid. Each grid point is represented by an initial condition, and its attractor label (left-tilt or right-tilt equilibrium) is determined by observing the simulated final angle where the disk eventually stops. Each of the 1600 samples is equally costly to label, but these samples make different contributions to estimating the BoA. Unlike a general classification problem, BoA are always contiguous, so if the number of attractors is known, it is sufficient to identify the boundaries between the basins; there is no need to check inside a basin for a region corresponding to a different attractor. For example, if there are two basins, there will be a single continuous boundary that defines the basins. However, in Fig.~\ref{fig:UFS}(c), only the 330 highlighted samples ($20\%$) were used to determine the boundary. In other words, uniform sampling wasted $80\%$ of its workload for unimportant samples. In order to address this inefficient sampling problem, this section proposes a novel sampling method which consists of three primary parts: additional sampling on the trajectories (AST), active learning (AL) and density-based sampling (DBS).


\subsection{Additional Sampling on Trajectories (AST) -- Get More Samples for ``Free''}
Unlike general classification problems where samples are independent, BoA estimation is able to take advantage of time series trajectories. On a trajectory, every state converges to the same attractor and thus shares the same label. In other words, when the label of an initial condition is determined by one numerical simulation or experiment, the generated time series trajectory can be sub-sampled to obtain additional labeled samples. 

\begin{figure}
	\centering
	\includegraphics[width=0.8\linewidth]{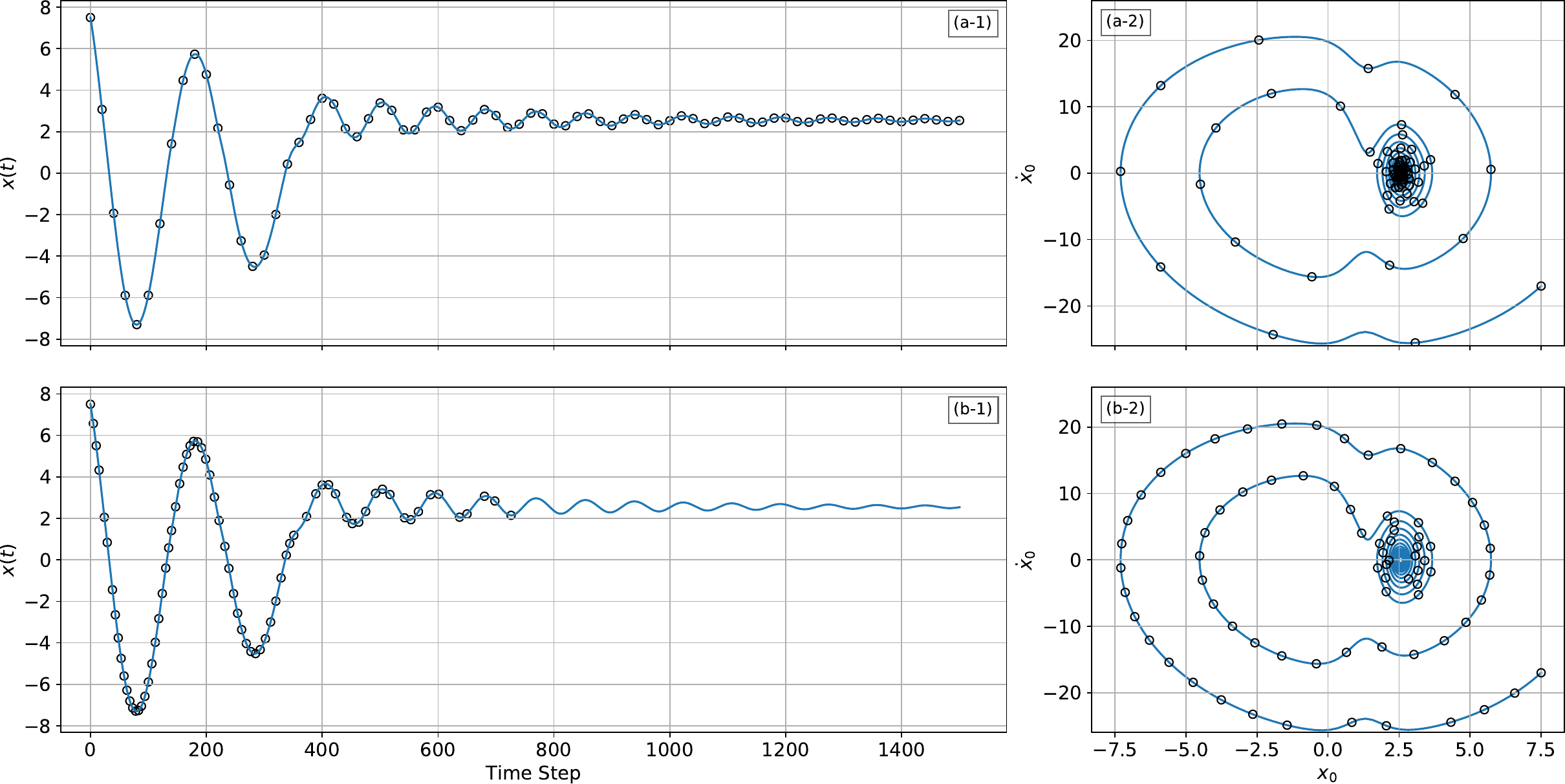}
	\caption{Additional sampling on trajectories (AST). (a-1) and (a-2) are sub-samples in the time domain and phase portrait respectively, by using a constant sampling step-size in the ``time'' domain. (b-1) and (b-2) are sub-samples in the time domain and phase portrait respectively, by using a constant sampling step-size in the ``state'' domain.}
	\label{fig:AST}
\end{figure}

For our method, the trajectory is sub-sampled by choosing samples that are a fixed distance apart in the state domain rather than the time domain. As shown in Fig.~\ref{fig:AST}(a-1, a-2), the sub-sampling with a constant step-size in the time domain results in the skewed distribution on the phase portrait. This distribution has an inaccurate description of BoA because many similar samples are gathering near the equilibrium while few samples contain the global BoA information. Therefore, as shown in Fig.~\ref{fig:AST}(b-1, b-2), the trajectory is sub-sampled using a constant step-size in the state domain, which leads to a more effective description of BoA globally.

\subsection{Active Learning (AL) -- Select ``Informative'' Samples}
Active learning (AL) is able to proactively select the unlabeled samples most likely to be ``informative'' and query an ``agent'' to obtain their labels. In the scenario of BoA estimation, the ``informative'' samples are the system's states near the BoA boundary, and the ``agent'' provides labels by determining the attractors where the system will eventually converge. The agent could be a numerical simulator if the governing equation of the system is known, or a human experimenter if the BoA for a real-world physical system is needed. Since the BoA estimation is essentially a classification problem and the informative samples are the states near a BoA boundary, our AL algorithm is built upon a support vector machine~(SVM) for its capability of providing a nonlinear classifier and easy interpretation of distances (also called ``margins'' in machine learning) from samples to the classifier's decision boundary~\cite{Balcan2007Margin}.

\begin{figure}
	\centering
	\includegraphics[width=0.8\linewidth]{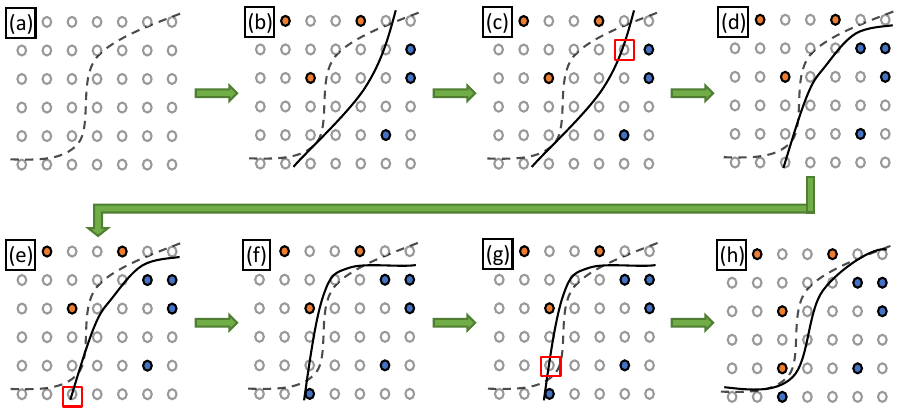}
	\caption{Margin-based active learning (margin-based AL). The grey dashed line is the real classification boundary while the black lines are the classifiers trained by the labeled samples. (a) Unlabeled data. (b) Label a random subset and fit a classifier. (c, e, g) Pick the closest point to the decision boundary. (d, f, h) Label the selected point and fit a new classifier. }
	\label{fig:AL}
\end{figure}

As shown in Fig.~\ref{fig:AL}, the margin-based AL needs to first generate a pool of unlabeled samples and randomly label a subset of them until different labels are observed. The following steps are then repeated until convergence: (1)~fit a SVM classifier using the labeled samples; (2)~label the unlabeled sample which has shortest distance (smallest ``margin'') to the current decision boundary. In the Fig.~\ref{fig:AL}, many of labeled samples are located near the real yet unknown classification boundary, which avoids wasted effort on labeling the less informative samples (e.g. the ones at the top-left and bottom-right corners).
%

When multiple samples have a similar distance from the boundary, the ``length'' of their trajectories can be used to break the tie, because longer trajectories usually provide more information. The ``length'' of a time series trajectory is defined as the total number of samples collected from the trajectory using a fixed sampling distance in the state domain. 

\begin{figure}
	\centering
	\includegraphics[width=0.8\linewidth]{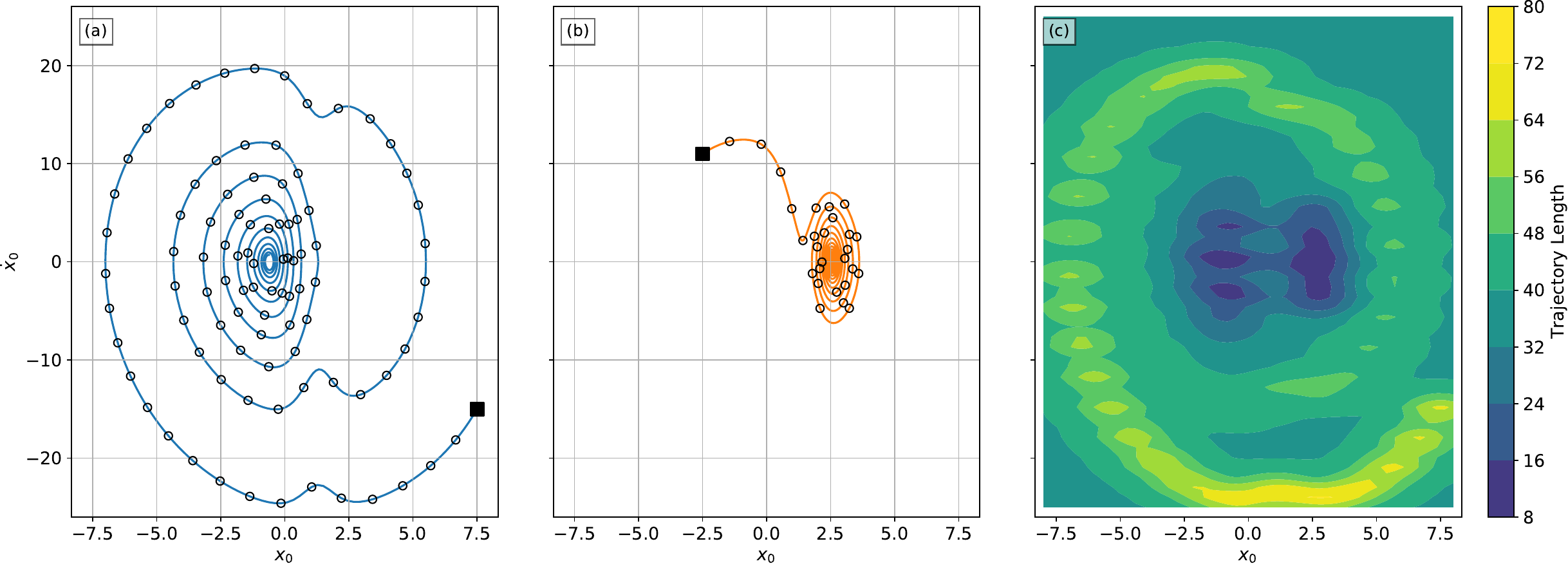}
	\caption{The margin-based AL gives two unlabeled samples (solid squares in (a) and (b)), which are assumed equally close to the decision boundary. The one in (a) generates a longer trajectory (83 samples) while the one in (b) generates a shorter trajectory (28 samples). (c) shows the prediction of trajectory length given a state by fitting the Gaussian process (GP) with the trajectories in (a) and (b).}
	\label{fig:trajectory}
\end{figure}

For the case illustrated in Fig.~\ref{fig:trajectory}(a, b), the margin-based AL determines the two unlabeled samples (solid squares) which are assumed equally close to the decision boundary. The one in (a) generates a longer trajectory and more labeled samples, so it has higher priority to be labeled. Since choosing between possible samples requires estimating the lengths prior to generating the trajectories, a predictive model for the trajectory length given a state is needed. A Gaussian process~(GP) model was selected since it can perform nonlinear regression, has controllable behavior when extrapolating, and needs no prior knowledge of the state distribution. Fig.~\ref{fig:trajectory}(c) shows a GP model which predicts the length of trajectories based on the samples in (a) and (b). In our sampling method, the GP model is updated every time a new trajectory is generated.

%
%
%

\subsection{Density-Based Sampling (DBS) -- Select ``Unfamiliar'' Samples for Exploration}
\label{sec:method_density}
Similar to the most learning algorithms, our sampling method should also deal with the exploration/exploitation dilemma. A sampling method built upon AL alone tends to exploit only and get stuck in the local BoA boundary.
In order to explore the entire state space and estimate the boundary globally, an auxiliary sampling approach based on the density of labeled samples is introduced. This density-based method prioritizes exploring the region in which the fewest samples have been labeled. 

\begin{figure}
	\centering
	\includegraphics[width=0.4\linewidth]{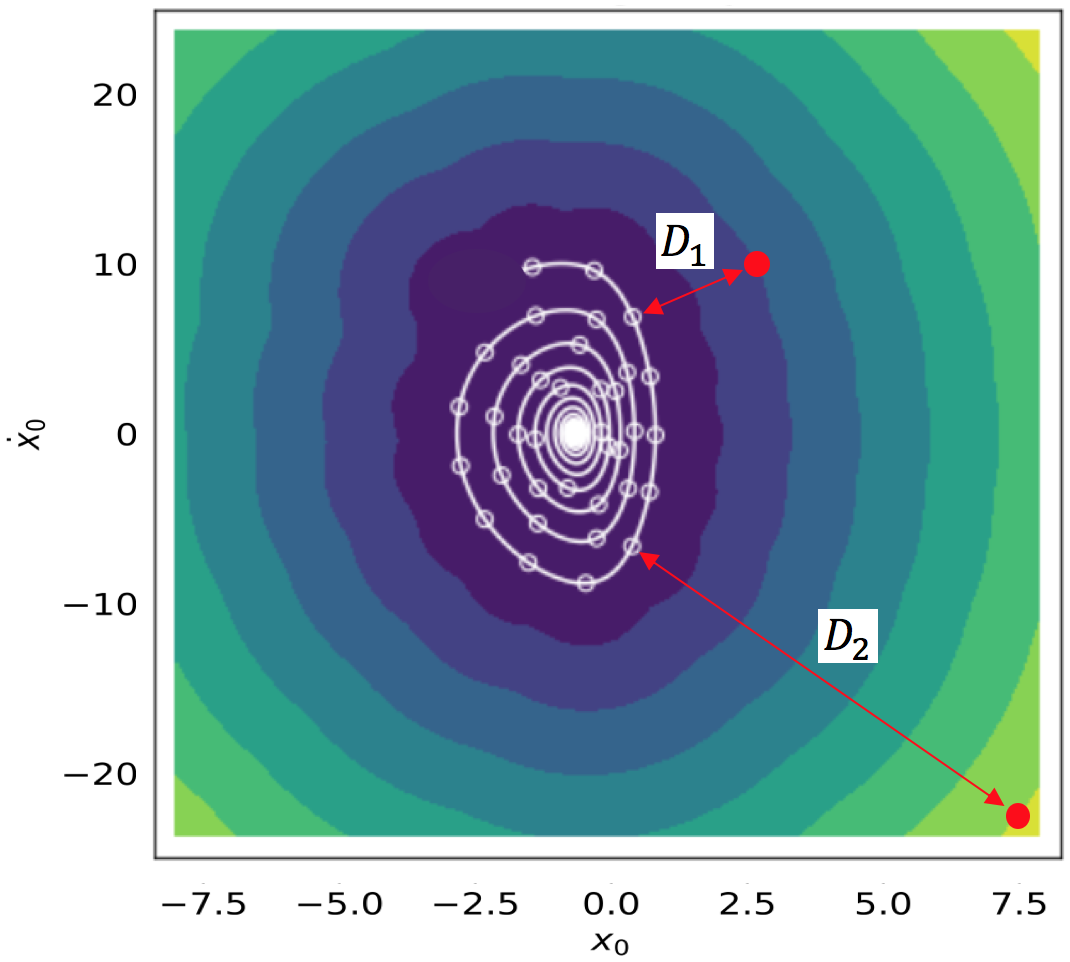}
	\caption{Density-Based Sampling (DBS). The white empty circles are labeled samples collected from a trajectory (white line), and the color in the heat map represents the density as well as the distance from a state point $\mathbf{X}$ to its closest labeled samples $D\left( \mathbf{X} \right)$: lighter color $\rightarrow$ larger $D\left( \mathbf{X} \right)$ $\rightarrow$ lower density. With regard to the unlabeled samples (red solid circles), $D_2 > D_1$ leads to that the right-bottom point has the priority to be sampled by DBS.}
	\label{fig:DBS}
\end{figure}

Inspired by the k-means++ algorithm which was proposed to spread out cluster centers~\cite{Arthur2006K}, we defined $D\left( \mathbf{X} \right)$ as the distance from an arbitrary state $\mathbf{X}$ to its closest ``labeled'' state. A larger $D\left( \mathbf{X} \right)$ indicates a lower density of labeled samples around the state $\mathbf{X}$, and $D\left( \mathbf{X} \right) = 0$ for a labeled state. For example, Fig.~\ref{fig:DBS} shows the density map upon which DBS selects an unlabeled sample. The white empty circles are the existing labeled samples. The central area has been sufficiently sampled, thus having a higher density than the areas near boarders. According to the definition of $D\left( \mathbf{X} \right)$ that indicates the density, the right-bottom state point has a larger distance $D_2$ than the other one $D_1$, thus having the priority to be selected by DBS. This density-based method selects the unlabeled sample with the largest $D\left( \mathbf{X} \right)$ and updates the distances in every sampling episode after new labeled samples are collected.

\subsection{Summary}
Integrating the three sections above (AST + AL + DBS) leads to a hybrid active learning (HAL) sampling method for estimating BoA (see Tab.~\ref{table:algr}). Several hyper-parameters need be predetermined: 1)~sampling distance on trajectories in AST; 2)~threshold $p$ of the top shortest distances in AL; 3)~the method of selecting between AL and DBS in each episode (random or alternate selection); and 4)~SVM kernel and GP kernel. There are also several tunable hyper-parameters inside the kernel function, but they could be automatically optimized during fitting.

It's worth noting that SVM was selected for its ability to quickly calculate distances from samples to the decision boundary in margin-based AL. However, once the sampling process is finished and calculating distances is no longer necessary, SVM might not be the best choice for the ultimate classifier, especially when the number of training samples is large. At the end of sampling process, the samples provide the necessary information to train other classifiers (such as k-nearest neighbors (KNN), random forests, neural networks, etc.) for better or faster estimation of the BoA.
\setcounter{magicrownumbers}{0}
\begin{table*}
	\caption{Hybrid active learning (HAL) sampling method for estimating basins of attraction (BoA)}
	\label{table:algr}
	\begin{tabular}{r|l}
		\hline
		\rownumber & Generate a pool of unlabeled samples (``states of initial conditions'')\\
		\rownumber & Randomly label a subset of unlabeled samples until different labels (``attractors'') are observed \\
		\rownumber & Obtain additional labeled samples from their time-series trajectories (\textbf{AST}) \\
		\rownumber & \textbf{for} episode = 1 : M \textbf{do} \\
		\rownumber & \quad Fit a support vector machine (SVM) classifier using the labeled samples \\
		\rownumber & \quad Fit a Gaussian process (GP) regressor using the length of trajectories generated \\
		\rownumber & \quad Randomly (or alternately) select one of the following sampling methods: \\
		\rownumber & \quad \textbf{(1)~Active Learning (AL):}\\
		\rownumber & \quad\quad Find the unlabeled samples within the top $p \times 100$\% shortest distance to the current SVM decision boundary \\
		\rownumber & \quad\quad Among them select the one with the longest trajectory length predicted by the GP regressor \\
		\rownumber & \quad \textbf{(2)~Density-Based Sampling (DBS):} \\
		\rownumber & \quad\quad Evaluate each unlabeled sample's distance to its closest labeled sample \\
		\rownumber & \quad\quad Among them select the one with largest distance \\
		\rownumber & \quad Label the selected sample and obtain additional labeled samples from its time series trajectory (\textbf{AST}) \\
		\rownumber & \textbf{end for} \\
		\hline
	\end{tabular}
\end{table*}
\section{Result}

\begin{figure}
	\centering
	\includegraphics[width=0.4\linewidth]{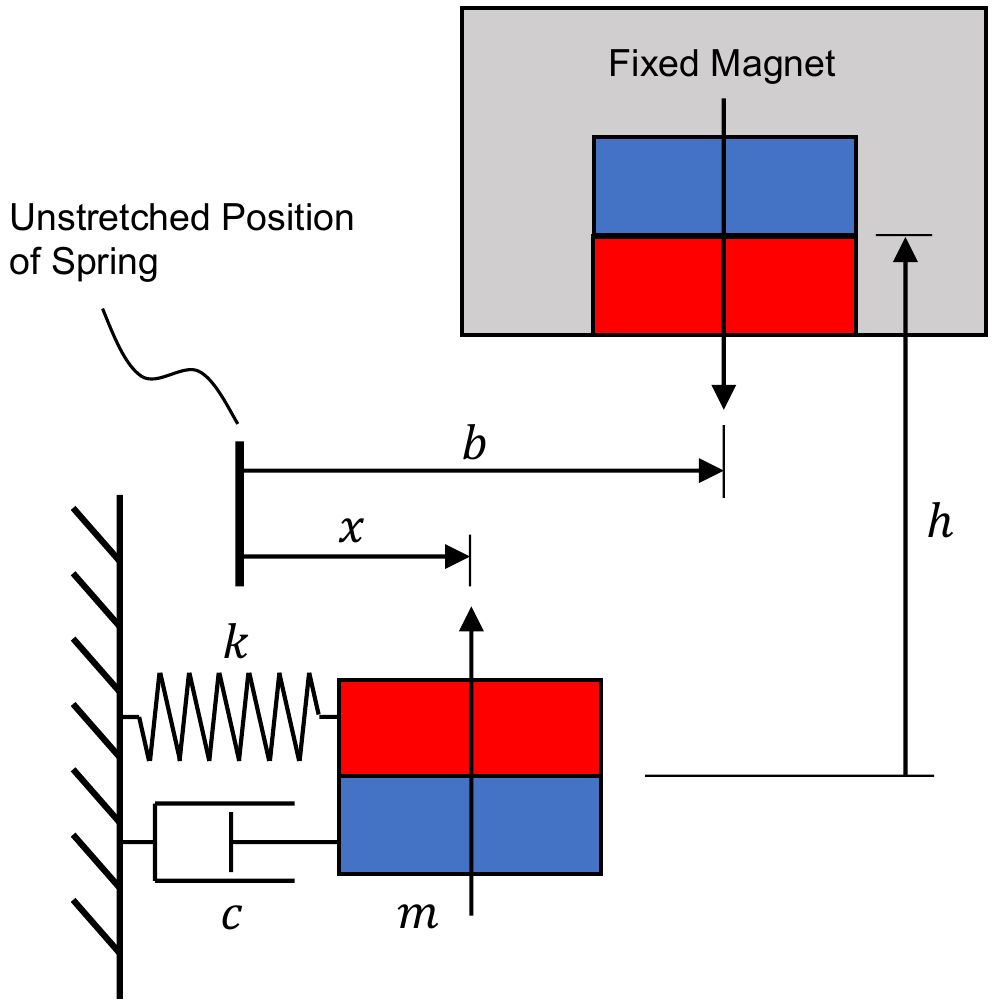}
	\caption{Schematic of the magnet-induced bistable system in Eq.~\eqref{eq:bistable_system}.}
	\label{fig:bistable_system}
\end{figure}

A magnet-induced bistable system in Fig.~\ref{fig:bistable_system} was used to illustrate the performance of our sampling method. It's worth noting that the governing equation provided below has no conflict with the essence of our model-free sampling method. This equation gives simulation results to represent the data collected from experiments, yet was not directly used for the BoA estimation. The governing equation of this bistable system was derived using the same method in Ref.~\cite{Wang2020Nonlinear} and can be written as:
\begin{equation}
	m\ddot x + c\dot x + k x = \alpha {\left( {x - b} \right)\left[ {12{h^2} - 3{{\left( {x - b} \right)}^2}} \right]} {{\left[ {{{\left( {x - b} \right)}^2} + {h^2}} \right]}^{-7/2}},
\label{eq:bistable_system}
\end{equation}
where $m=1$, $c=0.5$, $k=10$, $\alpha=100$, $h=1.5$, $b=1.3$. This bistable system has two stable equilibria (fixed-point attractors) $(x_\text{E}, \dot{x}_\text{E}) = (-0.612, 0)$ and $(2.555, 0)$ in the state domain of $x_0 \in [-8,\,8]$ and $\dot{x}_0 \in [-25,\,25]$. This 2-dimensional state domain was uniformly divided into a $50 \times 50$ grid, for 2500 unlabeled samples initially. For the hyper-parameters in the HAL sampling method, 1)~the sampling distance on trajectories in AST was set 0.07 after normalizing states to $[0,\,1]$; 2)~the threshold of the top shortest distances in AL was set $p = 0.05$; 3)~the sampling method alternated between AL and DBS per sampling episode. The Gaussian process (GP) predictor of the trajectory length and the support vector machine (SVM) classification for BoA boundary were both implemented by using scikit-learn library in Python. More specifically, GP was applied using GaussianProcessRegressor(alpha=1) and SVM was applied using SVC(kernel=`rbf'). For tuning the hyper-parameters inside the kernel function, GridSearchCV(scoring=``f1'', cv=5) was used with a parameter $5 \times 5$ grid of $C,\,\gamma \in \{ 0.01,\,0.1,\,1,\,10,\,100 \}$. 

\begin{figure}
	\centering
	\includegraphics[width=0.9\linewidth]{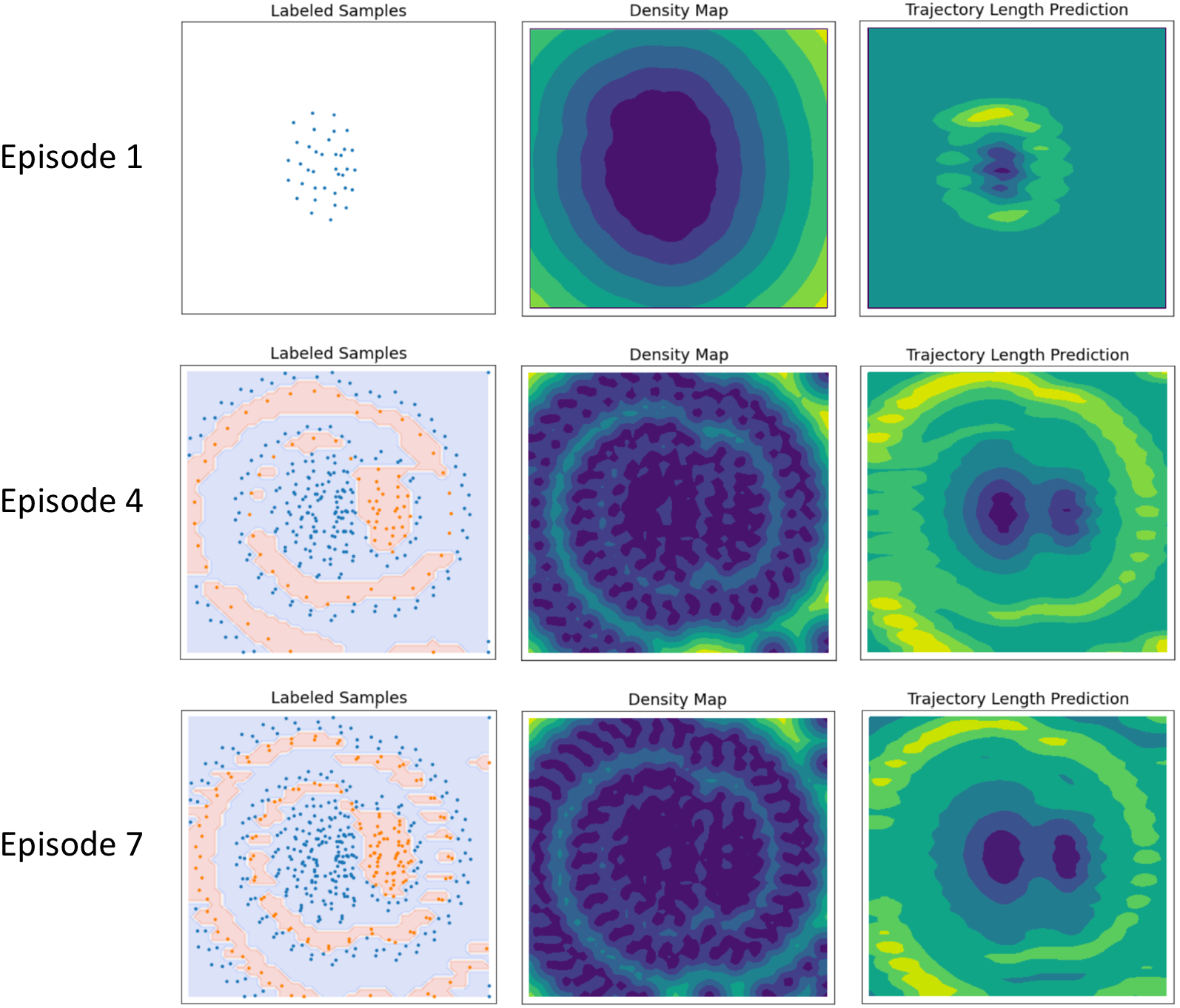}
	\caption{Sampling progress in the 1st, 4th and 7th episode. The first column is the labeled samples and BoA estimated by SVM (only exists after the second equilibrium was observed), the second column is the density map upon which DBS selected an unlabeled sample (lighter color indicates lower density of labeled samples), and the third column is the GP regressor for predicting the trajectory length (lighter color indicates a longer trajectory).}
	\label{fig:progress}
\end{figure}

\begin{figure}
	\centering
	\includegraphics[width=0.8\linewidth]{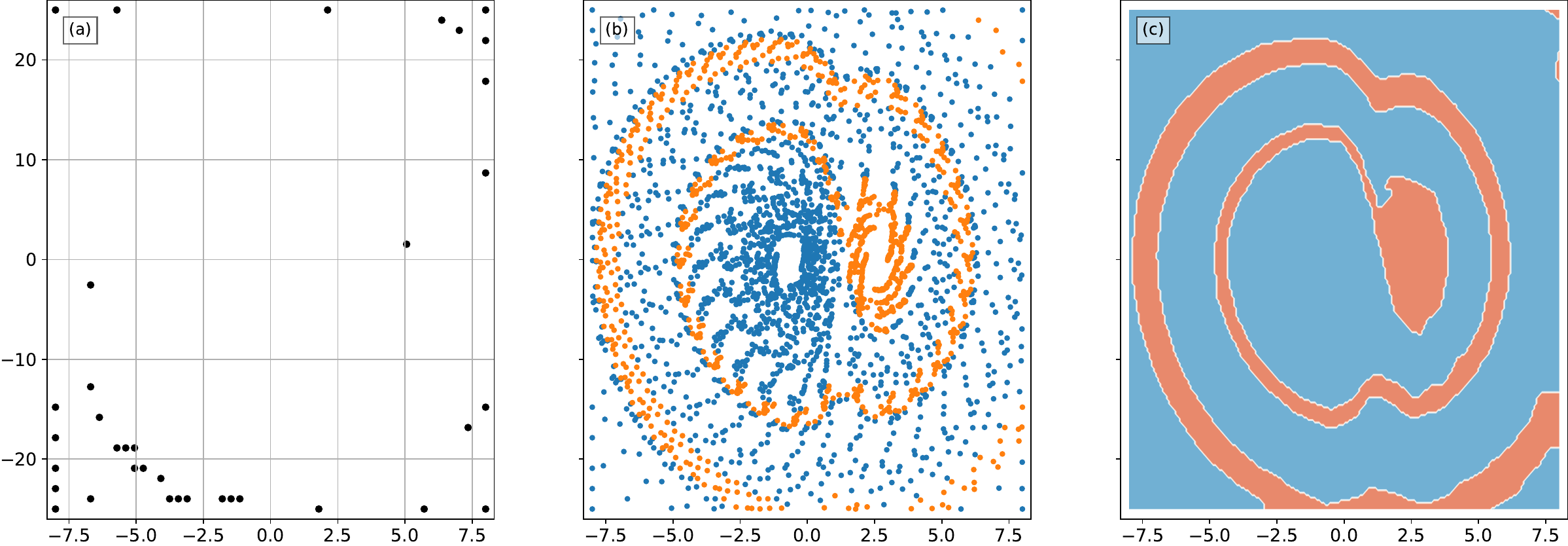}
	\caption{(a) The 36 samples selected to be labeled by our hybrid active learning (HAL) sampling method. (b) Additional samples on the time series trajectories generated by using samples in (a) as initial conditions. (c) BoA estimated by a neural network trained with the samples in (b).}
	\label{fig:showcase1}
\end{figure}

Figure~\ref{fig:progress} shows the sampling progress in the 1st, 4th and 7th episode. The density map and the GP regressor for the trajectory length were updated using all the samples that have been labeled until the present episode. Figure~\ref{fig:showcase1} gives the result of BoA estimation after 36 sampling episodes, where (a) shows the 36 samples selected by HAL to give labels, and (b) shows the additional labeled samples obtained from the time series trajectories starting from these 36 samples. A 3-layer neural network (input layer of 2 state variables -- 128 neurons -- ReLU -- 64 neurons -- ReLU -- 1 neuron -- sigmoid) was trained to give the BoA in Fig.~\ref{fig:showcase1}(c).

Given that the areas of basins are oftentimes imbalanced, ``F1-score'', which considers both recall and precision of a classifier, was selected to evaluate the estimated BoA. Tab.~\ref{tab:comparison} lists the minimum number of labels needed for different levels of F1-score for four sampling methods. The brute-force uniform sampling, as discussed in Sec.~\ref{sec:method}, wastes much effort on labeling the less informative samples that do not determine the BoA boundary, thus requiring the largest number of samples. Integrating uniform sampling with additional sampling from time series trajectories (uniform sampling + AST) gives little help for finding informative samples, but generates additional labeled samples every time one sample gets labeled, thus drastically reducing the sampling workload. The sampling method of AST + AL is capable of finding more informative samples near the boundary, which increases the sampling efficiency even more. However, as mentioned in Sec.~\ref{sec:method_density}, the sampling method built upon AL alone tends to get stuck in the local BoA boundary. It explains why this method behaves well for lower F1-scores, but leads to significantly low efficiency for higher F1-scores (our experiment stops after labeling 100 samples due to the unbearable computational time in AL). This problem was solved by introducing DBS, which explores global BoA boundaries. Combining all three sampling methods mentioned above (AST + AL + DBS) leads to our HAL sampling method, which shows a dominant advantage in the sampling efficiency in Tab.~\ref{tab:comparison}.
%
\begin{table}
	\centering
	\caption{Minimum number of labels needed for different levels of F1-score, a metric varying from 0 (bad estimation) to 1 (good estimation).}
	\label{tab:comparison}
	{
		\begin{tabular}{l|cccc}
			\hline
			\multicolumn{1}{c|}{Sampling Method} & \multicolumn{4}{c}{F1-Score} \\
			& 0.4 & 0.6 & 0.8 & 0.9  \\ \hline
			uniform sampling & 225 & 441 & 576 & 1225 \\
			uniform sampling + AST & 9 & 16 & 25 & 64 \\
			AST + AL & 4 & 7 & \textgreater 100 & \textgreater 100 \\
			AST + AL + DBS (our HAL method) & 4 & 7 & 15 & 35 \\ \hline
		\end{tabular}
	}
\end{table}

\section{Conclusion}
This paper introduces a hybrid active learning (HAL) sampling method for estimating a system's basins of attraction (BoA). The proposed method provides sufficient samples for a continuous map describing BoA, and more importantly, can be implemented in experiments where the system's underlying math model is unknown. It consists of three primary parts: 1)~an additional sampling of trajectories (AST) to maximize the number of samples obtained from each simulation or experiment; 2)~an active learning (AL) algorithm to exploit the local boundary of BoA; and 3)~a density-based sampling (DBS) method to explore the global boundary of BoA. An example of estimating BoA for a bistable nonlinear system was used to demonstrate the high efficiency of this HAL sampling method.

The author's believe future work should focus on advancing this approach to address current limitations.  More specifically, the current HAL sampling method is based on a binary classification approach, i.e. there are only two fixed-point attractors in the BoA. Although the method of ``one--vs--rest'' can be used for multi-label classification, this strategy leads to a workload that linearly increases with the number of attractors. A more efficient sampling method, which might be less affected by the number of attractors, is therefore needed. Second, this paper only illustrates the performance of the HAL sampling method on a 2-dimensional BoA that have smooth separatrix boundaries; its robustness and efficiency for estimating the BoA with higher dimensional and/or fractal boundaries provides additional opportunities for further investigation.

\bibliographystyle{unsrt}  
\bibliography{references}  

\begin{thebibliography}{10}

\bibitem{genesio1985estimation}
Roberto Genesio, Michele Tartaglia, and Antonio Vicino.
\newblock On the estimation of asymptotic stability regions: State of the art
  and new proposals.
\newblock {\em IEEE Transactions on automatic control}, 30(8):747--755, 1985.

\bibitem{chiang1988stability}
H-D Chiang, Morris~W Hirsch, and Felix~Fulih Wu.
\newblock Stability regions of nonlinear autonomous dynamical systems.
\newblock {\em IEEE Transactions on Automatic Control}, 33(1):16--27, 1988.

\bibitem{chiang1989stability}
H-D Chiang and James~S Thorp.
\newblock Stability regions of nonlinear dynamical systems: A constructive
  methodology.
\newblock {\em IEEE Transactions on Automatic Control}, 34(12):1229--1241,
  1989.

\bibitem{alberto2011characterization}
Luis Fernando~Costa Alberto and Hsiao-Dong Chiang.
\newblock Characterization of stability region for general autonomous nonlinear
  dynamical systems.
\newblock {\em IEEE transactions on automatic control}, 57(6):1564--1569, 2011.

\bibitem{margolis1963control}
S~Margolis and W~Vogt.
\newblock Control engineering applications of vi zubov's construction procedure
  for lyapunov functions.
\newblock {\em IEEE Transactions on Automatic Control}, 8(2):104--113, 1963.

\bibitem{rozgonyi2010determining}
Szabolcs Rozgonyi, Katalin Hangos, and G{\'a}bor Szederk{\'e}nyi.
\newblock Determining the domain of attraction of hybrid non-linear systems
  using maximal lyapunov functions.
\newblock {\em Kybernetika}, 46(1):19--37, 2010.

\bibitem{vannelli1985maximal}
Anthony Vannelli and Mathukumalli Vidyasagar.
\newblock Maximal lyapunov functions and domains of attraction for autonomous
  nonlinear systems.
\newblock {\em Automatica}, 21(1):69--80, 1985.

\bibitem{ohta2002piecewise}
Yuzo Ohta et~al.
\newblock Piecewise linear estimate of attractive regions for linear systems
  with saturating control.
\newblock In {\em Proc. of AMS}, 2002.

\bibitem{ohta1999stability}
Yuzo Ohta and Michio Onishi.
\newblock Stability analysis by using piecewise linear lyapunov functions.
\newblock {\em IFAC Proceedings Volumes}, 32(2):2083--2088, 1999.

\bibitem{balestrino2011logical}
Aldo Balestrino, Andrea Caiti, and Emanuele Crisostomi.
\newblock Logical composition of lyapunov functions.
\newblock {\em International journal of control}, 84(3):563--573, 2011.

\bibitem{Henrion2013Convex}
Didier Henrion and Milan Korda.
\newblock Convex computation of the region of attraction of polynomial control
  systems.
\newblock {\em IEEE Transactions on Automatic Control}, 59(2):297--312, 2013.

\bibitem{jarvis2003some}
Zachary Jarvis-Wloszek, Ryan Feeley, Weehong Tan, Kunpeng Sun, and Andrew
  Packard.
\newblock Some controls applications of sum of squares programming.
\newblock In {\em 42nd IEEE international conference on decision and control
  (IEEE Cat. No. 03CH37475)}, volume~5, pages 4676--4681. IEEE, 2003.

\bibitem{ichihara2009optimal}
Hiroyuki Ichihara.
\newblock Optimal control for polynomial systems using matrix sum of squares
  relaxations.
\newblock {\em IEEE Transactions on Automatic Control}, 54(5):1048--1053, 2009.

\bibitem{chesi2004computing}
Graziano Chesi.
\newblock Computing output feedback controllers to enlarge the domain of
  attraction in polynomial systems.
\newblock {\em IEEE Transactions on Automatic Control}, 49(10):1846--1853,
  2004.

\bibitem{chesi2004estimating}
Graziano Chesi.
\newblock Estimating the domain of attraction for uncertain polynomial systems.
\newblock {\em Automatica}, 40(11):1981--1986, 2004.

\bibitem{chesi2011domain}
Graziano Chesi.
\newblock {\em Domain of attraction: analysis and control via SOS programming},
  volume 415.
\newblock Springer Science \& Business Media, 2011.

\bibitem{chesi2005domain}
Graziano Chesi.
\newblock Domain of attraction: estimates for non-polynomial systems via lmis.
\newblock In {\em Proc. of 16th IFAC World Congress, Prague, Czech Republic},
  2005.

\bibitem{chesi2009estimating}
Graziano Chesi.
\newblock Estimating the domain of attraction for non-polynomial systems via
  lmi optimizations.
\newblock {\em Automatica}, 45(6):1536--1541, 2009.

\bibitem{hsu2013cell}
Chieh~Su Hsu.
\newblock {\em Cell-to-cell mapping: a method of global analysis for nonlinear
  systems}, volume~64.
\newblock Springer Science \& Business Media, 2013.

\bibitem{hsu1980unravelling}
CS~Hsu and RS~Guttalu.
\newblock An unravelling algorithm for global analysis of dynamical systems: An
  application of cell-to-cell mappings.
\newblock {\em J. Appl. Mech.}, 1980.

\bibitem{hsu1981generalized}
Chieh-Su Hsu.
\newblock A generalized theory of cell-to-cell mapping for nonlinear dynamical
  systems.
\newblock {\em J. Appl. Mech.}, 1981.

\bibitem{hsu1977determination}
CS~Hsu, HC~Yee, and WHt Cheng.
\newblock Determination of global regions of asymptotic stability for
  difference dynamical systems.
\newblock {\em J. Appl. Mech.}, 1977.

\bibitem{krauskopf1999two}
Bernd Krauskopf and Hinke Osinga.
\newblock Two-dimensional global manifolds of vector fields.
\newblock {\em Chaos: An Interdisciplinary Journal of Nonlinear Science},
  9(3):768--774, 1999.

\bibitem{osinga2010investigating}
Hinke Osinga, Bernd Krauskopf, Eusebius Doedel, and Pablo Aguirre.
\newblock Investigating the consequences of global bifurcations for
  two-dimensional invariant manifolds of vector fields.
\newblock {\em Discrete and Continuous Dynamical Systems}, 29(4):1309--1344,
  2010.

\bibitem{osinga2014computing}
Hinke~M Osinga.
\newblock Computing global invariant manifolds: Techniques and applications.
\newblock In {\em Proceedings of the International Congress of Mathematicians},
  volume~4, pages 1101--1123, 2014.

\bibitem{dellnitz1997subdivision}
Michael Dellnitz and Andreas Hohmann.
\newblock A subdivision algorithm for the computation of unstable manifolds and
  global attractors.
\newblock {\em Numerische Mathematik}, 75(3):293--317, 1997.

\bibitem{dellnitz2002set}
Michael Dellnitz and Oliver Junge.
\newblock Set oriented numerical methods for dynamical systems.
\newblock {\em Handbook of dynamical systems}, 2(1):900, 2002.

\bibitem{krauskopf2005survey}
Bernd Krauskopf, Hinke~M Osinga, Eusebius~J Doedel, Michael~E Henderson, John
  Guckenheimer, Alexander Vladimirsky, Michael Dellnitz, and Oliver Junge.
\newblock A survey of methods for computing (un) stable manifolds of vector
  fields.
\newblock {\em International Journal of Bifurcation and Chaos},
  15(03):763--791, 2005.

\bibitem{wang2020constrained}
Xue-She Wang, James~D Turner, and Brian~P Mann.
\newblock Constrained attractor selection using deep reinforcement learning.
\newblock {\em Journal of Vibration and Control}, page 1077546320930144, 2020.

\bibitem{Wang2018Dynamics}
Xue-She Wang, Michael~J. Mazzoleni, and Brian~P. Mann.
\newblock Dynamics of unforced and vertically forced rocking elliptical and
  semi-elliptical disks.
\newblock {\em Journal of Sound and Vibration}, 417:341 -- 358, 2018.

\bibitem{Wang2019Nonlinear}
Xue-She Wang and Brian~P. Mann.
\newblock Nonlinear dynamics of a non-contact translational-to-rotational
  magnetic transmission.
\newblock {\em Journal of Sound and Vibration}, 459:114861, 2019.

\bibitem{Balcan2007Margin}
Maria-Florina Balcan, Andrei Broder, and Tong Zhang.
\newblock Margin based active learning.
\newblock In {\em International Conference on Computational Learning Theory},
  pages 35--50. Springer, 2007.

\bibitem{Arthur2006K}
David Arthur and Sergei Vassilvitskii.
\newblock k-means++: The advantages of careful seeding.
\newblock Technical report, Stanford, 2006.

\bibitem{Wang2020Nonlinear}
Xue-She Wang and Brian~P. Mann.
\newblock Dynamics of a magnetically excited rotational system.
\newblock In Gaetan Kerschen, M.~R.~W. Brake, and Ludovic Renson, editors, {\em
  Nonlinear Structures and Systems, Volume 1}, pages 99--102, Cham, 2020.
  Springer International Publishing.

\end{thebibliography}

%

\end{document}